\documentclass[sigconf]{acmart}
\acmConference[Anonymous]{Anonymous Conference}{2025}{Anonymous Location}
\usepackage{multirow}
\usepackage{array}
\usepackage{pgfplots}
\pgfplotsset{compat=1.18}
\usepackage{tabularx}
\usepackage{caption}
\usepackage{pifont}
\usepackage{hyperref}
\newcommand{\xmark}{\ding{55}}
\begin{document}


\title{BaiJia: A Large-Scale Role-Playing Agent Corpus of Chinese Historical Characters}

\author{Ting Bai}
\email{baiting@bupt.edu.cn}
\affiliation{%
  \institution{Beijing University of Posts and Telecommunications}
  \city{Beijing}
  \country{China}
}
\author{Jiazheng Kang}
\email{kjz@bupt.edu.cn}
\affiliation{%
  \institution{Beijing University of Posts and Telecommunications}
  \city{Beijing}
  \country{China}
}
\author{Jiayang Fan}
\email{fjy01@bupt.edu.cn}
\affiliation{%
  \institution{Beijing University of Posts and Telecommunications}
  \city{Beijing}
  \country{China}
}


\renewcommand{\shortauthors}{Bai et al.}

\begin{abstract}  
We introduce a comprehensive large-scale role-playing agent corpus, termed \textbf{BaiJia}, that comprises various Chinese historical characters. This corpus is noteworthy for being the pioneering compilation of low-resource data that can be utilized in large language models (LLMs) to engage in AI-driven historical role-playing agents.
BaiJia addresses the challenges in terms of fragmented historical textual records in different forms and modalities, integrating various characters' information, including their biographical, literary, family relations, historical events, and so on. 
We conduct extensive experiments to demonstrate the effectiveness of our BaiJia agent corpus in bolstering the role-playing abilities of various foundational LLMs, and promoting the development and assessment of LLMs in the context of historical role-playing tasks. 
The agent corpus is available at \href{http://baijia.online}{baijia.online}\footnote{The evaluation benchmark is publicly available at \url{https://github.com/BAI-LAB/BaiJia}.}.
\end{abstract}

\maketitle
\section{Introduction}
Large language models (LLMs) show great potential to mimic human responses in role-playing research areas, enabling individuals to engage with historical characters in a lifelike and immersive manner. 
Equipping LLMs with role-playing capabilities provides a distinctive means of communicating with historical characters, fostering a deeper comprehension of their thoughts, actions, and the historical backgrounds of those who have made notable contributions to human history.

\begin{table}[htbp]
\centering
\caption{Dataset comparisons of role-playing agents.}
\small
\begin{tabular}{l|c|c|c}
\bottomrule
\textbf{Dataset} & \textbf{\# Characters} & \textbf{Source}           & \textbf{Low-Resource?} \\  
\hline \hline
ChatHaruhi       & 32                  & Anime, TV               & \xmark               \\  
InCharacter      & 32                  & Novels, Scripts         & \xmark               \\ 
CharacterEval    & 77                  & Novels, Scripts         & \xmark           \\ 
RoleLLM          & 100                 & Novels, Scripts          & \xmark               \\ \hline
\textbf{BaiJia} (Ours)    & 19,281       & History                 & \checkmark           \\ 
\bottomrule
\end{tabular}

\label{tab:dataset_comparison}
\end{table}

To empower LLMs with role-playing abilities, most existing studies, such as RoleLLM ~\cite{wang-etal-2024-rolellm}, InCharacter ~\cite{wang2024incharacter}, CharacterEval ~\cite{tu-etal-2024-charactereval}, and ChatHaruhi ~\cite{li2023chatharuhi}, have focused on Supervised Fine-Tuning (SFT) basic LLM models using collected or generated dialogues of characters. 
However, all of these approaches encounter the significant challenge of the high costs associated with data collection, which is a crucial resource in facilitating LLMs with role-playing capability.  
We summarize the data properties in existing role-playing studies, and highlight the differences of our BaiJia corpus in Table~\ref{tab:dataset_comparison}. 
We can see that most characters in existing studies are modern, anime, or fictional characters, there has been a notable lack of research dedicated for role-playing of historical characters. Building role-playing agents with historical characters raises great challenges from 
vast historical timelines they inhabit and the intricacies associated with the preservation of historical materials. 
Besides, the number of characters in existing research used for SFT role-playing agents is limited, which may hinder the model's capacity to fully realize its role-playing potential.


In this paper, we provide a low-resource data corpus, termed \textbf{BaiJia}, for role-playing agent construction. The information in our agent corpus spans numerous forms and modalities, including historical documents, ancient books, artworks, folklore, and oral traditions. BaiJia contains 19,281 Chinese historical characters from five dynasties, i.e., Tang, Song, Yuan, Ming, and Qing dynasties. It integrates different source information, including their biographical data, literary works, family relations, official positions, and historical events, providing a robust foundation for simulating the personalities, behaviors, and dialogues of these characters.
To the best of our knowledge, we are the first to construct a large-scale role-playing agent corpus for Chinese historical characters. Our contributions are as follows:

\begin{table*}
    \centering
    \small
    \caption{ The resume template of Chinese historical characters. We present an example resume of a famous poet \emph{"Li Bai"}. \emph{Completion} shows the proportion of characters for whom we have collected this type of information.}
    \begin{tabular}{c|c|c|c|c|c} \bottomrule 
        \textbf{Category} & \textbf{Sub-category} & \textbf{Content} & \textbf{\#Type} & \textbf{Example} & \textbf{Completion} \\ \hline   \hline 
        \multirow{7}{*}{\textbf{Profile}} 
            & Basic Information & Name, Dynasty, Birth, Death, Age, Family Division & 19,281 & Li Bai (701–762 CE)[...] & 100\% \\ \cline{2-6}
            & Alias Information & Alias, Alias Type & 19 & Taibai & 51\% \\ \cline{2-6}
            & Social Division & Describing Social Division & 199 & Poet & 36\% \\ \cline{2-6}
            & Personal Introduction & A Brief Biographical Overview & 19062 & Li Bai is a poet[...] & 98\% \\ \cline{2-6}
            & Geographic Information & Start Year, End Year, Geographic Type & 21 & Birthplace Is Rencheng & 69\% \\ \cline{2-6}
            & Wealth Information & Location, Actions, Description, Quantity & 3 & - & 1\% \\ \cline{2-6}
            & Event Information & Event Name, Role in Events, Event Description & 250 & An Lushan Rebellion & 2\% \\ \hline  
        \multirow{2}{*}{\textbf{Relations}} 
            & Family Relations & Detailed Family Relations & 297 & Li Bai's father is Li Ke. & 44\% \\ \cline{2-6}
            & Other Relations & All relationships except for family & 424 & friend is Meng Haoran. & 35\% \\ \hline  
        \multirow{3}{*}{\textbf{Career}} 
            & Entry Information & Entry Type, Entry Age, Entry Year & 175 & Official Recruitment & 43\% \\ \cline{2-6}
            & Appointment Information & Official Position, Start Year, End Year, Proxy, Location & 3,905 & Hanlin Scholar & 53\% \\ \cline{2-6}
            & Institutional Information & Institution name, Role of the Institution & 26 & Hanlin Academy & 1\% \\ \hline
        \multirow{3}{*}{\textbf{Achievement}} 
            & Literary Writings & Title, Year, Content, Role of writings & 17,714 & <<Li Taibai>> & 23\% \\ \cline{2-6}
            & Poetry\&Essay & Title, Author, Content & 310,754 & Silent thoughts & 98\% \\ \cline{2-6}
            & Dialogue Content & Type, Location, Background, Dialogue & 192,810 & - & 100\% \\  \bottomrule
    \end{tabular}
    \label{tab:combined_character_profile}
\end{table*}

\begin{itemize}
\item  We contribute a large-scale Chinese historical character agent corpus termed BaiJia, which firstly collects low-resource data for LLMs to conduct AI-driven historical role-playing.
\item We design comprehensive evaluation dimensions and release an evaluation benchmark for the role-playing task of historical characters. 
\item  We conduct extensive experiments to show the usefulness of our agent corpus in improving the role-playing capability of different basic LLMs. 
\end{itemize}

\section{Dataset Construction}
\begin{figure}[htbp]
    \centering
    \includegraphics[width=0.47\textwidth]{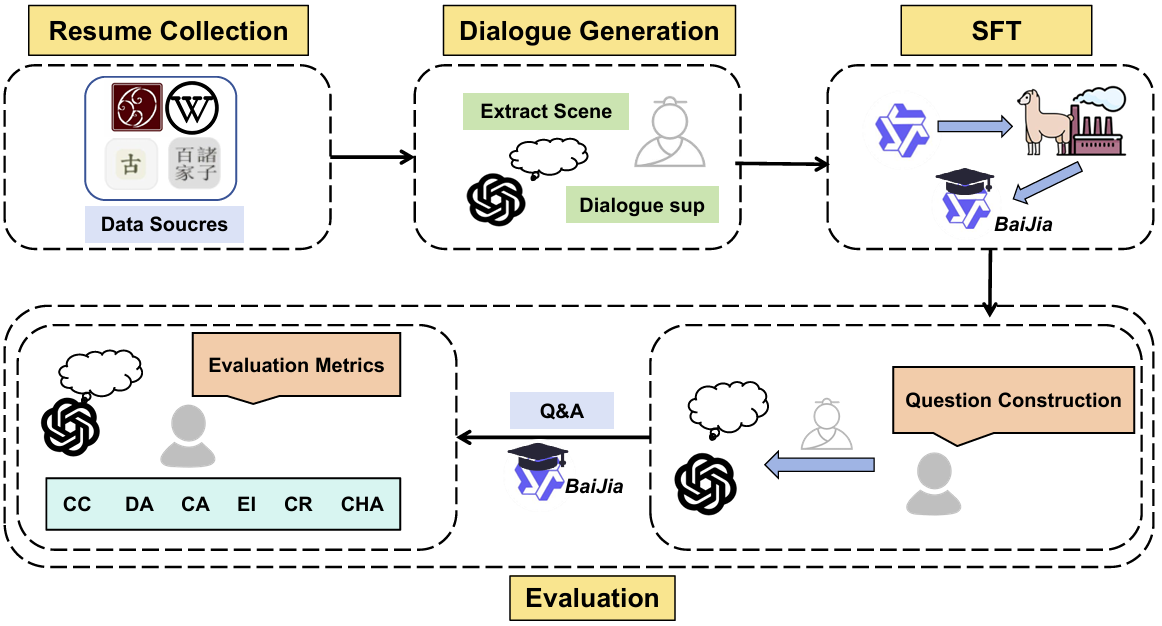} 
    \caption{The pipeline of role-playing agent construction.}
    \label{fig}
\end{figure}

The pipeline for the construction and evaluation of role-playing agents is shown in Fig.~\ref{fig}. We highlight the key steps of data construction, dialogue generation, and model evaluation process.
The data corpus that we contributed for role-playing agent construction includes three parts: the collection of character resumes, the generation of dialogues that were used to fine-tune LLMs, and the construction of questions that were used to evaluate the role-playing capability of LLMs.

\subsection{Resume Collection}
We collect diverse character information from multiple sources, e.g., CBDB\footnote{\href{https://projects.iq.harvard.edu/cbdb}{https://projects.iq.harvard.edu/cbdb}}, Wikipedia\footnote{\url{https://en.wikipedia.org/wiki}}, \emph{Gushiwen} website\footnote{\url{https://www.gushiwen.cn/}}, to construct Informative and comprehensive character resume in role-playing agent construction. 
The characters are from five major dynasties in Chinese history: 3,020 characters from \emph{Tang}, 5,964 characters from \emph{Song}, 972 characters from  \emph{Yuan}, 4,564 characters from  \emph{Ming}, and  4,761 characters from \emph{Qing} dynasties. 
The resumes of each character consist of information about their profiles, their relationships, and their work. The detailed introduction of the resume template is shown in Table~\ref{tab:combined_character_profile}, which can be summarized into 15 sub-categories that cover the basic profile, relations, career, and achievement of characters. 


\subsection{Dialogue Generation}
After constructing the resumes of characters, we generate the dialogues that are used for SFT LLMs. Following Character-LLM~\cite{shao2023character}, we adopt a two-step dialogue generation approach: (1) extracting the character's dialogue scenes: we adopt GPT-4o-mini to extract 10 unique scenes based on the resume of each character. These scenes include palace dialogues, family conversations, and literary debates. The prompts focus on the character's social relations, life events, and works, ensuring an authentic historical setting; (2) generating dialogues related to these scenes: under the background of different scenes, we use GPT-4o-mini to automatically generate the questions and simulate responses that align with the historical context of characters.

Finally, we utilize the LLaMA-Factory framework~\cite{zheng2024llamafactory} to perform LoRA fine-tuning on LLMs with resumes and the generated dialogue information of characters. This process enables the LLMs to acquire the ability for role-playing.

\begin{table*}[htbp]
\small
\centering
\setlength{\tabcolsep}{3pt} 
\caption{Performance comparisons of different LLMs. The results of LLMs with our agent corpus are marked with underline.}
\label{tab:result_reordered}
\begin{tabular}{c|ccc|ccc|ccc|ccc|ccc|ccc|c}
\bottomrule
\textbf{Model} 
& \multicolumn{3}{c|}{\textbf{CC}} 
& \multicolumn{3}{c|}{\textbf{DA}} 
& \multicolumn{3}{c|}{\textbf{CA}} 
& \multicolumn{3}{c|}{\textbf{EI}} 
& \multicolumn{3}{c|}{\textbf{CR}} 
& \multicolumn{3}{c|}{\textbf{CHA}} 
& \textbf{Avg.Imp} $\uparrow$ \\
\hline \hline
\textbf{ChatGLM3-6B} 
 & 3.66 & \underline{4.52} & \textbf{+23.5\%} 
 & 3.64 & \underline{4.10} & \textbf{+12.6\%} 
 & 3.50 & \underline{3.90} & \textbf{+11.4\%} 
 & 3.57 & \underline{3.90} & \textbf{+9.3\%} 
 & 3.15 & \underline{3.75} & \textbf{+19.0\%} 
 & 3.68 & \underline{4.48} & \textbf{+21.7\%} 
 & \textbf{+16.3\%} \\ 
\textbf{Qwen2.5-7B} 
 & 4.10 & \underline{4.63} & \textbf{+13.0\%} 
 & 4.20 & \underline{4.32} & \textbf{+2.9\%} 
 & 4.04 & \underline{4.22} & \textbf{+4.5\%} 
 & 4.07 & \underline{4.12} & \textbf{+1.2\%} 
 & 3.94 & \underline{4.09} & \textbf{+3.8\%} 
 & 4.39 & \underline{4.70} & \textbf{+7.1\%} 
 & \textbf{+5.42\%} \\ 
\textbf{Llama-3.1-8B} 
 & 3.58 & \underline{4.50} & \textbf{+25.7\%} 
 & 3.63 & \underline{4.00} & \textbf{+10.2\%} 
 & 3.60 & \underline{3.91} & \textbf{+8.6\%} 
 & 3.69 & \underline{3.92} & \textbf{+6.2\%} 
 & 3.37 & \underline{3.78} & \textbf{+12.0\%} 
 & 3.47 & \underline{4.34} & \textbf{+25.3\%} 
 & \textbf{+14.7\%} \\ 
\hline
\textbf{Llama-3.1-70B} 
 & 4.00 & \underline{4.70} & \textbf{+17.5\%} 
 & 3.94 & \underline{4.21} & \textbf{+6.9\%} 
 & 3.82 & \underline{4.01} & \textbf{+5.0\%} 
 & 3.84 & \underline{3.99} & \textbf{+3.9\%} 
 & 3.60 & \underline{3.91} & \textbf{+8.6\%} 
 & 3.93 & \underline{4.60} & \textbf{+17.5\%} 
 & \textbf{+9.9\%} \\ 
\textbf{Qwen2.5-72B} 
 & 4.30 & \underline{4.78} & \textbf{+11.2\%} 
 & 4.48 & \underline{4.66} & \textbf{+4.0\%} 
 & 4.17 & \underline{4.28} & \textbf{+2.6\%} 
 & 4.19 & \underline{4.28} & \textbf{+2.1\%} 
 & 4.14 & \underline{4.25} & \textbf{+2.7\%} 
 & 4.83 & \underline{4.96} & \textbf{+2.7\%} 
 & \textbf{+4.2\%} \\ 
\textbf{DeepSeekV2.5} 
 & 4.03 & \underline{4.92} & \textbf{+22.1\%} 
 & 4.11 & \underline{4.54} & \textbf{+10.6\%} 
 & 4.01 & \underline{4.19} & \textbf{+4.5\%} 
 & 3.88 & \underline{4.17} & \textbf{+7.5\%} 
 & 3.97 & \underline{4.14} & \textbf{+4.4\%} 
 & 4.15 & \underline{4.90} & \textbf{+18.1\%} 
 & \textbf{+11.2\%} \\ 
\hline
\textbf{Baichuan-NPC} 
 & 4.04 & \underline{4.45} & \textbf{+10.1\%} 
 & 3.71 & \underline{4.00} & \textbf{+7.8\%} 
 & 3.56 & \underline{3.85} & \textbf{+8.1\%} 
 & 3.56 & \underline{3.83} & \textbf{+7.6\%} 
 & 3.30 & \underline{3.70} & \textbf{+12.1\%} 
 & 4.11 & \underline{4.46} & \textbf{+8.5\%} 
 & \textbf{+9.0\%} \\ 
\textbf{Xingchen} 
 & 3.29 & \underline{4.26} & \textbf{+29.5\%} 
 & 3.30 & \underline{3.94} & \textbf{+19.4\%} 
 & 3.08 & \underline{3.72} & \textbf{+20.8\%} 
 & 3.18 & \underline{3.73} & \textbf{+17.3\%} 
 & 2.74 & \underline{3.52} & \textbf{+28.5\%} 
 & 3.38 & \underline{4.29} & \textbf{+26.9\%} 
 & \textbf{+23.7\%} \\ 
\bottomrule
\end{tabular}
\end{table*}

\subsection{Question Construction}
To evaluate the usefulness of our character agent corpus, we construct a question dataset that is used for historical role-playing agent evaluation.
For each character, we construct 15 questions from five thematic aspects:
\emph{Personal Background}, \emph{Era Background}, \emph{Family \& Social Connections}, \emph{Thoughts, Personality \& Values}, \emph{Achievements \& Contributions}.
We use GPT-4o-mini to generate knowledge-oriented questions, ensuring that the questions do not directly reveal specific information. For example, instead of asking, “Your hometown is Yong’an; how did it influence you?” we phrase it as, “Where is your hometown? How did it influence you?” This approach ensures that questions remain open-ended, allowing for a more accurate assessment of the model’s ability to acquire and understand character knowledge in the development of role-playing agents of role-playing. 

\section{Experiment}

\subsection{Experimental Setup}
\subsubsection{Baseline Models} 
To verify the usefulness of our constructed agent corpus, we conduct experiments on different kinds of LLMs, including the general LLMs: i.e., ChatGLM~\cite{glm2024chatglm}, Qwen\footnote{\url{https://qwenlm.github.io/blog/qwen2.5/}}, Lama~\cite{grattafiori2024llama3herdmodels}, DeepSeek~\cite{deepseekv2}, and the role-playing LLMs (RP-LLM): i.e.,  BaiChuanNPC\footnote{\url{https://npc.baichuan-ai.com/}}.
 and Tongyi Xingchen\footnote{\url{https://tongyi.aliyun.com/xingchen/}}.
Our LLM BaiJia has been fine-tuned on Qwen2.5-7B with resume and dialogue information from our constructed agent corpus, ensuring our LLM remains lightweight and highly specialized. 
Table~\ref{tab:baseline_models} summarizes the difference among them according to the size of parameters and applications. 


\begin{table}[htbp]
    \centering
    \small
     \caption{The comparison of different types of LLMs}
    \label{tab:baseline_models}
    \begin{tabular}{cccc}
        \toprule
        \textbf{Model} & \textbf{\# Parameter} & \textbf{RP-LLM} & \textbf{Open Source} \\ \hline
        \midrule
        ChatGlm3-6B & 6B & \xmark    & \checkmark \\
        Qwen2.5-7B & 7B & \xmark & \checkmark \\
        Llama-3.1-8B & 8B & \xmark & \checkmark \\  \hline 
        Llama-3.1-70B & 70B & \xmark & \checkmark \\
        Qwen2.5-72B & 72B & \xmark & \checkmark \\ 
        DeepSeekV2.5 & 236B & \xmark & \checkmark \\
        \hline 
        Xingchen & -- & \checkmark & \xmark \\
        Baichuan-NPC & -- & \checkmark & \xmark \\ 
        \textbf{BaiJia}(Ours)  & 7B & \checkmark & partially \\
        \bottomrule
    \end{tabular}
\end{table}

\subsubsection{Evaluation Metrics} 
We design a comprehensive evaluation benchmark to assess the capabilities of LLMs on six dimensions: \emph{Character Consistency} (CC), \emph{Dialogue Ability} (DA), \emph{Character Appeal} (CA), \emph{Emotional Expression \& Intellectual Depth} (EI), \emph{Creativity \& Role Depth Expansion} (CR), and \emph{Cultural \& Historical Appropriateness} (CHA). 
Except for the evaluation dimensions, i.e., CC, DA and CA, from existing role-playing benchmarks~\cite{wang-etal-2024-rolellm,li2023chatharuhi,wang2024incharacter,tu-etal-2024-charactereval}, we propose three new dimensions, i.e., EI, CR, and CHA, that are specifically designed to evaluate the deep-level spiritual aspect of historical characters, including their emotion, creation and culture understanding. 
Each of these six dimensions contains two sub-dimensions, forming a comprehensive assessment of role-playing performance with a total of twelve sub-dimensions.
Detailed evaluation dimensions are outlined in Table~\ref{tab:character_analysis}.

\begin{table}[h!]
\centering
\small
\caption{Evaluation metrics of role-playing agent.}
\renewcommand{\arraystretch}{1.2} 
\begin{tabular}{c|c}
\toprule
\textbf{Metrics} & \textbf{Sub-dimension} \\ \hline \hline
\multirow{2}{*}{\textbf{CC}} & Alignment with Character Background \\
 & Alignment with Dynasty Background \\ \hline
 \multirow{2}{*}{\textbf{DA}} & Coherence and Logicality of Dialogue \\
 & Interactivity in Dialogue \\ \hline
\multirow{2}{*}{\textbf{CA}} & Charm and Attractiveness of the Character \\
 & Emotional Impact in Character Interactions \\ \hline
\multirow{2}{*}{\textbf{EI}} & Authentic Emotional Expression \\
 & Intellectual Depth and Philosophical Views \\ \hline
\multirow{2}{*}{\textbf{CR}} & Character Personality and Innovative Thinking \\
 & Further Development of Character Complexity \\ \hline
\multirow{2}{*}{\textbf{CHA}} & Language Style Matching Character Identity and Era \\
 & Alignment with Historical and Cultural Views of the Era \\ \bottomrule
\end{tabular}
\label{tab:character_analysis}
\end{table}

\begin{figure*}[htbp]
    \centering
    \includegraphics[width=0.91\textwidth]{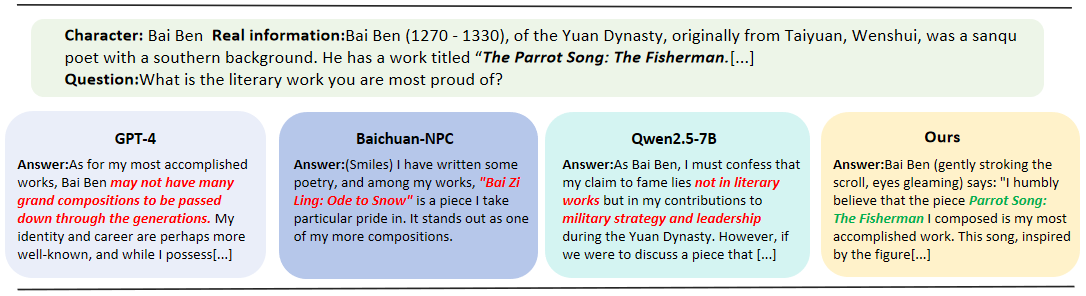}
\caption{Comparison of responses from different LLMs for the question to character \emph{Bai Ben}. According to his resume, we highlight the correct answers in green color. The red color indicates the fabricated answer or false answers.}
    \label{fig:case_study}
\end{figure*}

\subsection{Experimental Results}

To evaluate the effectiveness of our agent corpus, we compare the performance of different LLMs with and without the incorporation of our corpus. The results are shown in Table~\ref{tab:result_reordered}. We can see that: (1) After incorporating the information of character resumes (results marked with underline), the role-playing capabilities of all kinds of LLMs gain significant improvements over six evaluation dimensions; (2) Despite the increased capabilities of advanced LLMs, such as Qwen2.5-72B v.s. Qwen2.5-7B and Llama-3.1-70B v.s. Llama-3.1-8B, the enhancements achieved through our data corpus still be significant. This demonstrates that our corpus effectively fills the data gaps of current LLMs in role-playing tasks; (3) For LLMs specialized for role-playing applications, such as BaichuanNPC and Xingchen, we observe that they are unable to portray Chinese historical characters effectively. This may be attributed to the limited availability and widespread distribution of historical data.
(4) The greatest improvements achieved in the dimensions of Character Consistency (CC) and Culture \& Historical Appropriateness (CHA), showing the powerfulness of our agent corpus in assisting LLMs to generate contextually coherent and realistic dialogues.

\begin{figure}[htbp]
    \centering
    \begin{minipage}[b]{\linewidth}
        \centering
        \includegraphics[width=0.9\linewidth]{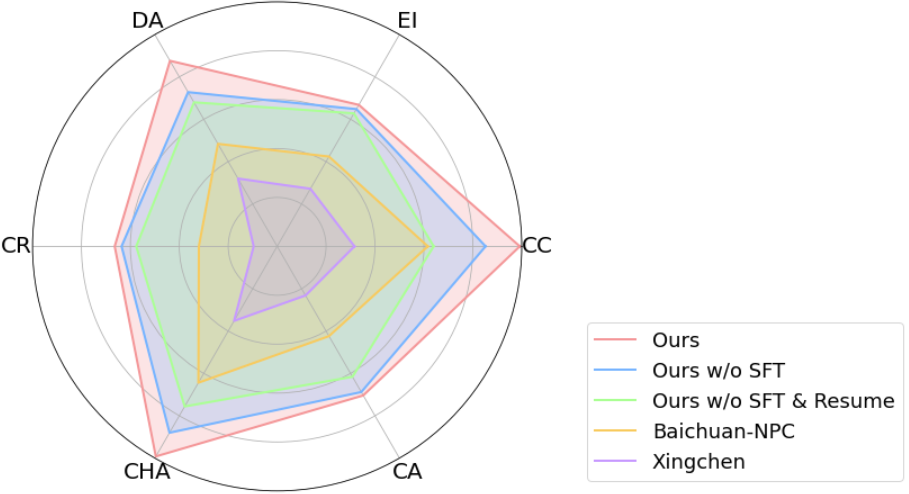}  
         \caption{Radar chart shows the performance of the fully optimized LLM ("Ours") and its variants across six evaluation dimensions.}
        \label{fig:ambition_study}
    \end{minipage}%
\end{figure}

\subsection{Experimental Analysis}
\subsubsection{Ablation Study}
An ablation study is conducted to evaluate the effects of our agent corpus, i.e., the resume information and the generated dialogue information of characters.   
Our model, i.e.,\emph{Ours}, integrates resumes and is SFT with dialogue information based on the Qwen2.5-7B framework. We compare it with its degradation versions, i.e., \emph{w/o SFT}: utilizing resume information only, and \emph{w/o SFT \& Resume}: neither conduct SFT with dialogue information nor incorporate resume information.  
As shown in Figure~\ref{fig:ambition_study}, we can see that the fully optimized LLM "\emph{Ours}" achieves superior performance across all evaluation dimensions. Without SFT or resume information, it leads to noticeable performance degradation, showing the usefulness of our corpus in enhancing the consistency and comprehensive abilities of role-playing LLMs.
Compared to LLMs that are specialized for role-playing agents, without the data resources, they even perform worse than the general LLMs due to their fine-tuning for distinctive kinds of specific characters from Anime or Novels.

\subsubsection{Case Study}
To intuitively show the effectiveness of our agent corpus, we present a case study of a historical character \emph{Bai Ben} and compare the responses generated from various LLMs. 
As shown in Fig.~\ref{fig:case_study}, for the question \emph{"What is the literary work you are most proud of?"},
Baichuan-NPC generates a title of work  ( i.e., \emph{"Ode to Snow"}) in the responses, but it is fictional. GPT-4 and Qwen2.5-7B are unable to provide responses, i.e., \emph{"may not have many grand compositions to be passed down"} from GPT-4 and \emph{"not in literary works"} from Qwen2.5-7B, owing to their deficiency in relevant knowledge.
After incorporating our agent corpus, the fine-tuned Qwen2.5-7B accurately responds \emph{"Parrot Song: The Fisherman"} as Bai Ben's most accomplished work. This response aligns with historical records and demonstrates the superiority of our corpus in capturing and reproducing historical character information. 

\section{Conclusion}
We contribute a high-quality agent corpus of Chinese historical characters, which is vital to improving the role-playing capability of large language models. 
By aggregating fragmented information from diverse data sources and integrating missing data, we collect an invaluable data resource in the realm of historical role-playing agents. This large-scale agent corpus is a groundbreaking contribution to low-resource historical AI role-playing research.

\bibliographystyle{ACM-Reference-Format}
\bibliography{Reference}
\end{document}